\documentclass[11pt,a4paper]{article}
\usepackage{times,latexsym}
\usepackage{url}
\usepackage{amsmath}
\usepackage{amsfonts}
\usepackage[T1]{fontenc}
\usepackage{array} 
\usepackage{authblk}
\usepackage{arydshln} 
\usepackage{pifont}
\usepackage[dvipsnames]{xcolor}
\usepackage{bbding}
\usepackage[whole]{bxcjkjatype}
\makeatletter
\newcommand\email[2][]%
   {\newaffiltrue\let\AB@blk@and\AB@pand
      \if\relax#1\relax\def\AB@note{\AB@thenote}\else\def\AB@note{\relax}%
        \setcounter{Maxaffil}{0}\fi
      \begingroup
        \let\protect\@unexpandable@protect
        \def\thanks{\protect\thanks}\def\footnote{\protect\footnote}%
        \@temptokena=\expandafter{\AB@authors}%
        {\def\\{\protect\\\protect\Affilfont}\xdef\AB@temp{#2}}%
         \xdef\AB@authors{\the\@temptokena\AB@las\AB@au@str
         \protect\\[\affilsep]\protect\Affilfont\AB@temp}%
         \gdef\AB@las{}\gdef\AB@au@str{}%
        {\def\\{, \ignorespaces}\xdef\AB@temp{#2}}%
        \@temptokena=\expandafter{\AB@affillist}%
        \xdef\AB@affillist{\the\@temptokena \AB@affilsep
          \AB@affilnote{}\protect\Affilfont\AB@temp}%
      \endgroup
       \let\AB@affilsep\AB@affilsepx
}
\makeatother
\newcommand\blfootnote[1]{%
  \begingroup
  \renewcommand\thefootnote{}\footnote{#1}%
  \addtocounter{footnote}{-1}%
  \endgroup
}
\usepackage[acceptedWithA]{tacl2021v1}
\usepackage{booktabs}
\usepackage{graphics}
\usepackage{graphicx}
\usepackage{multirow}
\usepackage{CJKutf8}
\usepackage{amssymb}
\usepackage{footmisc}

\newcommand{\code}[1]{\texttt{#1}}
\usepackage{xspace,mfirstuc,tabulary}

\newif\iftaclinstructions
\taclinstructionsfalse 
\iftaclinstructions
\renewcommand{\confidential}{}
\renewcommand{\anonsubtext}{(No author info supplied here, for consistency with
TACL-submission anonymization requirements)}
\newcommand{\instr}
\fi

\iftaclpubformat 

\else

\fi
\newcommand{\rev}[1]{{\color{blue} #1}}
\renewcommand{\rev}[1]{{#1}}

\title{Minimum Bayes Risk Decoding for Error Span Detection in Reference-Free Automatic Machine Translation Evaluation}

\author[1]{\vspace{0mm}Boxuan Lyu}
\author[2]{Haiyue Song}
\author[3]{Hidetaka Kamigaito}
\author[2]{Chenchen Ding}
\author[2]{\\ Hideki Tanaka}
\author[2]{Masao Utiyama}
\author[1]{Kotaro Funakoshi}
\author[1]{Manabu Okumura\vspace{-4mm}}
\affil[1]{\small \vspace{-1mm}Institute of Science Tokyo}
\affil[2]{\small \vspace{-1mm}National Institute of Information and Communications Technology}
\affil[3]{\small \vspace{-1mm}Nara Institute of Science and Technology}
\email{\vspace{-1mm}\small\url{{lyu,funakoshi,oku}@lr.first.iir.isct.ac.jp}}
\email{\vspace{-1mm}\small\url{{haiyue.song,chenchen.ding,hideki.tanaka,mutiyama}@nict.go.jp}}
\email{\vspace{-1mm}\small\url{kamigaito.h@is.naist.jp}}
\date{}
\newcommand{\lyu}[1]{#1}
\begin{document}
\maketitle
\begin{abstract}
Error Span Detection (ESD) extends automatic machine translation (MT) evaluation by localizing translation errors and labeling their severity.
Current generative ESD methods typically use Maximum a Posteriori (MAP) decoding, assuming that the model-estimated probabilities are perfectly correlated with similarity to the human annotation, but we often observe a higher likelihood assigned to an incorrect annotation than to the human one.
In contrast, we apply Minimum Bayes Risk (MBR) decoding to generative ESD.
We use a sentence- or span-level similarity function for MBR decoding, which selects candidate hypotheses based on their approximate similarity to the human annotation.
Experimental results on the WMT24 Metrics Shared Task show that MBR decoding significantly improves span-level performance and generally matches or outperforms \rev{both the MAP and Majority Voting baselines} at the system and sentence levels.
To reduce the computational cost of MBR decoding, we further distill its decisions into a model decoded via greedy search, removing the inference-time latency bottleneck.\blfootnote{This  work was done during the first author's internship at National Institute of Information and Communications Technology, Kyoto, Japan.}\footnote{Our models and code are available at: \url{https://github.com/vlaks425/MBR-ESD}}
\end{abstract}


\section{Introduction}
\label{sec:intro}
Automatic metrics for machine translation (MT) provide a fast proxy for costly human evaluation.
From early n-gram metrics such as BLEU (\citealp{bleu}) to learned metrics such as COMET (\citealp{comet22}), correlations with human judgments have improved at the system and sentence levels.
However, most metrics still output a single scalar score per sentence (\citealp{wmtm24}), hiding diagnostic information such as error location and severity, which are crucial for model analysis and post-editing.

Error Span Detection (ESD), \rev{also referred to as Word-level Quality Estimation (WQE) in some studies}, extends MT evaluation by asking metrics not only to score a translation but also to identify error spans and their severities.
ESD has been part of the WMT Quality Estimation shared task since 2023 (\citealp{wmtqe23}).
Current ESD methods fall into two paradigms: token-classification approaches, such as xCOMET (\citealp{xcomet}), which treat ESD as per-token labeling, and generative approaches, such as GEMBA-MQM (\citealp{gemba_mqm}), which prompt large language models (LLMs) to output MQM-style (\citealp{mqm}) error spans in zero- or few-shot settings.


Current state-of-the-art LLM-based generative ESD methods (\citealp{wmtm24}) typically rely on Maximum a Posteriori (MAP) decoding. 
The underlying assumption of MAP decoding is that the model-estimated probabilities are perfectly correlated with similarity to the human annotation. 
However, this assumption does not always hold.
\rev{We} observe instances where the model log-likelihood is higher for an incorrect annotation produced by MAP decoding than for the correct human annotation, signifying a model error; \rev{we present a concrete example of this in Table~\ref{tab:logp}.}
This discrepancy, consistent with findings in other text generation tasks (\citealp{mbr_nmt2}; \citealp{brio}), leads us to question the optimality of MAP as a decoding objective for ESD.

\begin{table*}[!t]
\setlength\tabcolsep{12pt}
\centering
\scalebox{1.0}{
\begin{tabular}{ccc}
\toprule
System & Error Span Annotation & Log-Likelihood\\
\midrule
Human&Ich wollte fliegen, \textcolor{red}{da} ich ein Kind war. &-10.84\\
Generative Model (with MBR)&Ich wollte fliegen, \textcolor{red}{da ich ein Kind war}.&-5.99\\
Generative Model (with MAP)&\textcolor{red}{Ich wollte fliegen}, \textcolor{red}{da ich ein Kind war}.&\textbf{-2.86}\\
\bottomrule
\end{tabular}}
\caption{A failure case for MAP decoding on the WMT24 Metrics Shared Task (English$\rightarrow$German) using \code{Llama-3.3-70B-Inst}. The source is ``I've wanted to fly since I was a child.'' \rev{The reference translation is ``Ich wollte fliegen, seit ich ein Kind war.''} Error spans are highlighted in red. The log-likelihood of the correct human annotation is lower than the incorrect MAP decoding output.}
\label{tab:logp}
\end{table*}

Minimum Bayes Risk (MBR) decoding is a reranking method widely studied in MT (\citealp{mbr_smt1}; \citealp{mbr_smt2}) and speech recognition (\citealp{mbr_speech1}; \citealp{mbr_speech2}).
In this work, we apply MBR decoding to the ESD model with the objective of maximizing utility against the human annotation, independent of the model-estimated probabilities.
Since the human annotation is unavailable at test time, we approximate it with a set of model-generated ``support hypotheses'' and select the hypothesis with the highest expected utility against this set. 

We define the utility function as a similarity function over hypotheses pairs and consider both sentence-level and span-level variants.
The proposed sentence-level utility function measures similarity based on the difference between the severity-weighted error span counts of the two hypotheses.
For the span-level utility function, we \rev{adopt} the official $\textsc{F1}(\cdot,\cdot)$ metric from the WMT QE Shared Task (\citealp{wmtqe23}), \rev{as well as $\textsc{SoftF1}(\cdot,\cdot)$, an improved utility function that we propose to address a defect in $\textsc{F1}(\cdot,\cdot)$.}

Experimental results on the WMT24 Metrics Shared Task show that our MBR decoding, when instantiated with the proposed $\textsc{SoftF1}(\cdot,\cdot)$ utility function and applied with sufficiently large candidate sets, significantly improves span-level performance and generally matches or outperforms \rev{both the MAP and Majority Voting baselines} at the system and sentence levels.
To avoid the high computational cost of MBR decoding, we further distill its decisions into a model decoded via greedy search, removing the inference-time latency bottleneck.


\section{Related Work}
\label{sec:related_work}
To our knowledge, our work is the first to apply reranking techniques to MT metrics.

\subsection{MT Automatic Metrics}
Early MT metrics such as BLEU (\citealp{bleu}), METEOR (\citealp{meteor}), and chrF (\citealp{chrf}) rely on surface-level matching and output a single scalar score per sentence.
Neural metrics, including BertScore (\citealp{BERTScore:}), BLEURT (\citealp{bleurt}), COMET (\citealp{comet}), and MetricX (\citealp{metricx24}), show higher correlations with human judgments (\citealp{wmtm22}; \citealp{wmtm23}; \citealp{wmtm24}) but usually still produce sentence-level scores only, limiting their usefulness for applications such as post-editing.

As human evaluation has shifted from scalar scores (e.g., Direct Assessment) to multidimensional annotations such as MQM (\citealp{mqm}), \rev{evaluation has moved toward finer-grained, error-localizing assessment.
This task is known as ESD (also referred to as WQE); it underpins downstream uses such as AI-assisted human evaluation (\citealp{wqe_ai_eval}) and human post-editing (\citealp{qe4pe}).
Approaches to it span a spectrum: unsupervised methods that read error signals from the internal states of translation models (\citealp{unsup_qe}; \citealp{xcomet_calibrating}), supervised token-classification metrics such as xCOMET (\citealp{xcomet}), and generative methods such as GEMBA-MQM (\citealp{gemba_mqm}) and AutoMQM (\citealp{automqm}).
Among these, LLM-based generative approaches have attracted growing attention for their flexibility---often requiring only prompt changes---and their competitive, often state-of-the-art performance (\citealp{gen_metric1}; \citealp{gen_metric2}; \citealp{gen_metric3}; \citealp{gen_metric4}; \citealp{gen_metric5}; \citealp{wmtm24}).}

\subsection{Evaluation Metrics for ESD}
\label{sec:metrics4esd}
The alignment between model-predicted ESD annotations and the human annotation can be evaluated at the system, sentence, and span levels. 

Soft Pairwise Accuracy (SPA) (\citealp{spa}) and Pairwise Accuracy with Tie Calibration (${\text{Acc}}_{\text{eq}}^{*}$) (\citealp{acc_eq}) measure ranking consistency.
Specifically, SPA operates at the system level by evaluating pairwise rankings of different MT systems based on aggregated scores, whereas ${\text{Acc}}_{\text{eq}}^{*}$ functions at the sentence level, assessing the relative ordering of individual translation sentence pairs.

At the span level, to our knowledge, the only existing metric for ESD is the official metric of the WMT QE Shared Task (\citealp{wmtqe23}), $\textsc{F1}(\cdot, \cdot)$, which computes character-level precision and recall between a hypothesis and a reference annotation.
This metric can also be used as an MBR utility function when the reference is a support hypothesis.
\rev{However, as we detail in \S\ref{sec:mbr_f1}, $\textsc{F1}(\cdot,\cdot)$ has a critical defect in its handling of empty annotations, which motivates the soft alternative $\textsc{SoftF1}(\cdot,\cdot)$ that we propose in \S\ref{sec:softf1}.}

\subsection{MBR Decoding}
\label{sec:mbr_related_work}
MBR decoding has a long history in speech recognition (\citealp{mbr_speech1}; \citealp{mbr_speech2}), word alignment (\citealp{mbr_word_align}), and statistical MT (\citealp{mbr_smt1}; \citealp{mbr_smt2}). 
Recently, several studies have revisited MBR decoding for text generation, demonstrating promising results (\citealp{llm_mbr1}; \citealp{llm_mbr2}; \citealp{smbr}). 
These studies suggest that MBR decoding can help overcome some of the limitations inherent in MAP. 
Furthermore, the properties (\citealp{mbr_properties1}; \citealp{mbr_properties2}) and efficiency (\citealp{mbr_faster1}; \citealp{mbr_faster2}; \citealp{mbr_faster3}) of MBR decoding remain active areas of research.

\section{MAP and MBR Decoding}
In this section, we describe MAP and MBR decoding in the context of general conditional generative models.
Given an input $x$, a conditional generative model $M(\cdot|x)$ defines a probability distribution over the hypothesis space.
Decoding for this model can be viewed as consisting of two phases: \emph{hypothesis generation} and \emph{decision}. 
In the hypothesis generation phase, a procedure such as sampling or beam search is used to produce $N$ candidate hypotheses $\mathcal{C}=\{h_0,\ldots,h_{N-1}\}$ from the model $M(\cdot\mid x)$.
In the decision phase, a decision rule assigns a score ${score}_{h}$ to each $h\in\mathcal{C}$ and returns $\mathop{\mathrm{argmax}}\limits_{h \in \mathcal{C}}score_{h}$ as the final output.

\subsection{MAP Decoding}
Formally, let $\mathcal{H}$ denote the entire hypothesis space. MAP decoding selects the hypothesis with the highest model-estimated probability from $\mathcal{H}$:
\begin{align}
h^{\mathit{MAP}}  &= \mathop{\mathrm{argmax}}\limits_{h \in \mathcal{H}}score^{\mathit{MAP}}_{h},\nonumber\\
score^{\mathit{MAP}}_{h}
&= M(h\mid x).\nonumber
\end{align}
\label{eq:map}
Finding the global maximizer over $\mathcal{H}$ is typically intractable due to the enormous size of the space.
In practice, MAP decoding is approximated by restricting the search to a finite set of candidate hypotheses $\mathcal{C} \subset \mathcal{H}$:
\begin{align}
h^{\mathrm{MAP}} \approx \mathop{\mathrm{argmax}}\limits_{h \in \mathcal{C}}M(h\mid x).\nonumber
\end{align}

\subsection{MBR Decoding}
Unlike MAP, MBR explicitly optimizes task utility rather than model-estimated probability. 
Let $\mathcal{Y}$ denote the space of the (unknown) ground-truth output $y$ for input $x$ and $u(\cdot,\cdot)$ be a utility function,   
MBR decoding selects the hypothesis that \emph{minimizes the expected risk}---or equivalently, \emph{maximizes the expected utility}---with respect to $\mathcal{Y}$:
\begin{align}
h^{\mathit{MBR}} & = \mathop{\mathrm{argmax}}\limits_{h_c \in \mathcal{C}} 
score^{\mathit{MBR}}_{h},\nonumber\\
score^{\mathit{MBR}}_{h} & = {\mathbb{E}_{y \in \mathcal{Y}}\left[u(h, y) \mid x\right]}.
\label{def:mbr}
\end{align}
Since $y$ is unknown in practice, MBR decoding approximates this expectation using a set of ``support hypotheses'' $\mathcal{S}\subseteq\mathcal{H}$ drawn from $M(\cdot|x)$:
\begin{align}
\label{def:mbr_approx}
score^{\mathit{MBR}}_{h} &\approx {\mathbb{E}_{h_s \in \mathcal{S}}\left[u(h, h_s) \mid x\right]},\\
&\approx \sum_{h_s \in \mathcal{S}} u(h, h_s) {M}(h_s\mid x).\nonumber
\end{align}
A common practice is to use the same set for candidate and support hypotheses ($\mathcal{C}=\mathcal{S}$) and to assume a uniform distribution over support hypotheses, yielding a simple average utility:
\begin{align}
score^{\mathit{MBR}}_{h} &\approx \frac{1}{|\mathcal{S}|} \sum_{h_s \in \mathcal{S}} u(h_c, h_s).
\label{def:mbr_appro2}
\end{align}

\section{Error Span Detection with MAP Decoding}
\subsection{Generative ESD Task}\label{sec:ESD}
We focus on the ESD task in the reference-free setting: given an input $x=(s,t)$ consisting of a source sentence $s$ and its translation $t$, the model outputs an error annotation (a set of error spans with associated severities).

Let $E$ denote a hypothesis containing a finite number of major and minor error spans, and let $\mathcal{H}_{\text{ESD}}$ represent the space of all admissible hypotheses. 
We denote the distribution of the (unknown) ground-truth human annotation as $\mathcal{Y}_{\text{ESD}}$. 
The generative ESD model, $M_{\text{ESD}}(E|x)$, estimates the probability of a hypothesis $E \in \mathcal{H}_{\text{ESD}}$.

For notation, we use $E_{c}$ to denote a candidate hypothesis, $E_{s}$ for a support hypothesis, and $E_{y}$ for the human annotation.
Similarly, $\mathcal{C}_{\text{ESD}}$ and $\mathcal{S}_{\text{ESD}}$ represent the sets of candidate and support hypotheses, respectively.

\subsection{MAP Decoding for ESD}
\label{sec:map_for_esd}
Under MAP decoding, the decision score for a hypothesis $E \in \mathcal{C}_{\text{ESD}}$ is defined as its model-estimated probability:
\begin{align}
score^{\mathit{MAP}}_{E} = M_{\text{ESD}}(E\mid x).\nonumber
\end{align}
\label{eq:map_e}
As discussed in \S\ref{sec:intro}, the MAP objective implicitly assumes that model-estimated probabilities are perfectly correlated with similarity to the human annotation. 
\rev{However, as illustrated by the example in Table~\ref{tab:logp} (\S\ref{sec:num_spans}), this assumption frequently does not hold for ESD.}
This discrepancy motivates the adoption of MBR decoding as a robust alternative.

\section{Proposed Method}

\subsection{MBR Decoding for ESD}
\begin{figure}[!h]
\centering
     \includegraphics[width=0.95\linewidth]{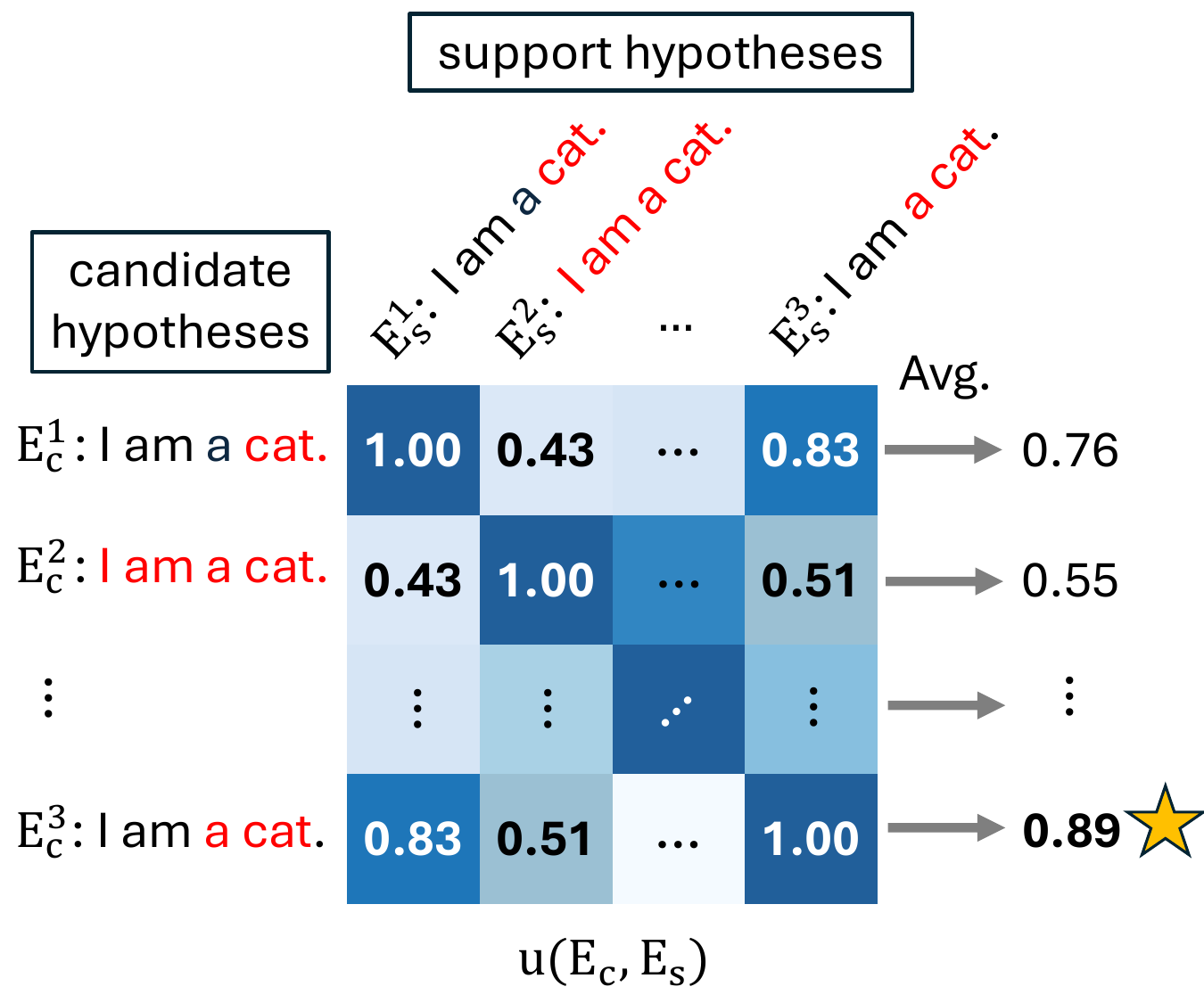}
     \caption{An overview of our MBR decoding for generative ESD models. Error spans are highlighted in red.}
    \label{fig:overview}
\end{figure}
We propose applying MBR decoding to generative ESD models to address the limitations of MAP decoding. 
Figure \ref{fig:overview} provides an overview of our method.

Analogous to Eq.~\ref{def:mbr}, the MBR score for a hypothesis $E$ is defined as:
\begin{align}
score^{\mathit{MBR}}_{E} & = {\mathbb{E}_{E_y \in \mathcal{Y}_{\text{ESD}}}\left[u(E, E_y) \mid x\right]}.
\label{def:mbr_esd}
\end{align}
As in Eq.~\ref{def:mbr_appro2}, we approximate this expectation using a finite sample $\mathcal{S}_{\mathrm{ESD}}\subset \mathcal{H}_{\mathrm{ESD}}$:
\begin{align}
score^{\mathit{MBR}}_{E} &\approx \frac{1}{|\mathcal{S}_{\text{ESD}}|} \sum_{E_s \in \mathcal{S}_{\text{ESD}}} u(E, E_s).
\label{def:mbr_esd_appro}
\end{align}
We specify the utility functions $u(\cdot,\cdot)$ below.

\subsection{Utility Functions of MBR Decoding in ESD}
\label{sec:mbr_utility}
In this section, we describe three instantiations of MBR decoding for generative ESD models. 
The utility functions fall into two categories: sentence and span level.

Notably, although SPA and ${\text{Acc}}_{\text{eq}}^{*}$ (mentioned in \S\ref{sec:metrics4esd}) are standard metrics for ESD, they cannot serve as MBR utility functions. 
This is because they measure ranking consistency over a dataset rather than computing a similarity score for a single pair of error annotations.

\subsubsection{MBR-$\textsc{ScoreSim}$}
\label{sec:sent_mbr}

We introduce \textbf{MBR-$\textsc{ScoreSim}$}, which employs our proposed $\textsc{ScoreSim}(\cdot,\cdot)$ as the utility function. 
$\textsc{ScoreSim}(\cdot,\cdot)$ is a sentence-level utility function that measures the similarity between two hypotheses based on severity-weighted error span counts.

Let $N^{\text{maj}}(\cdot)$ and $N^{\text{min}}(\cdot)$ denote the counts of major and minor error spans in $E$, respectively. 
Following common practice (\citealp{mqm}; \citealp{metricx24}), we first convert the error spans into a scalar sentence-level score using the standard MQM scoring function:
\begin{align}
\scalebox{0.8}{$\textsc{Score}(E) = \max(w_{\text{maj}} N^{\text{maj}}(E)+w_{\text{min}} N^{\text{min}}(E), \alpha).$}
\label{def:score}
\end{align}
Here, $w_{\text{maj}}$ and $w_{\text{min}}$ are negative severity weights ($w < 0$). 
These negative values reflect the subtractive nature of the MQM protocol (\citealp{mqm}), which penalizes the accumulation of error spans rather than rewarding quality. 
Finally, $\alpha < 0$ defines a lower bound on the total penalty.

We then define the similarity between $E_c$ and $E_s$ as one minus the normalized absolute difference between their scores, using $|\alpha|$ for normalization to yield a value in $[0, 1]$:
\begin{align}
\scalebox{1.0}{$\textsc{ScoreSim}(E_c,E_s)=1-\frac{|\,\textsc{Score}(E_c)-\textsc{Score}(E_s)\,|}{|\alpha|}.$}\nonumber
\end{align}

\subsubsection{MBR-$\textsc{F1}$}
\label{sec:mbr_f1}
For span-level utility, we \rev{adopt} \textbf{MBR-$\textsc{F1}$}, which employs the standard $\textsc{F1}(\cdot, \cdot)$ metric from the WMT QE Shared Task. 
This function calculates the harmonic mean of character-level precision and recall between two annotations (see Appendix A.1 for details).

While $\textsc{F1}(\cdot, \cdot)$ is widely used, \rev{we identify a critical defect in its handling of empty annotations.}
Specifically, if the support hypothesis $E_s$ is empty (i.e., contains no errors), $\textsc{F1}(\cdot, E_s)$ assigns the minimum utility ($0$) to any non-empty candidate $E_c$, regardless of how minor the error in $E_c$ might be (see Appendix A.1 for the derivation).
This ``all-or-nothing'' behavior is problematic for two reasons.
\rev{First, as an evaluation metric, it fails to provide a soft penalty when the human annotation is error-free}, ignoring the nuance that a hypothesis with a single minor error is arguably better---and more similar to an error-free translation---than one with multiple major errors.
\rev{Second, as an MBR utility function, it assigns minimum utility to any non-empty candidate whenever the support hypothesis is empty.
Given the diversity of support hypotheses (\citealp{mbr_properties2}) and the growing prevalence of perfect translations (\citealp{wmtm24}), cases with empty human annotations or empty support hypotheses are likely to be frequent.
We therefore regard this behavior as a significant limitation, which motivates our soft alternative, $\textsc{SoftF1}(\cdot,\cdot)$, introduced in \S\ref{sec:softf1}.}

\subsubsection{MBR-$\textsc{SoftF1}$}
\label{sec:softf1}
We introduce \textbf{MBR-$\textsc{SoftF1}$}, an instance of MBR decoding using our proposed $\textsc{SoftF1}(\cdot,\cdot)$ as the utility function.
$\textsc{SoftF1}(\cdot,\cdot)$ provides a continuous, ``soft'' evaluation of mismatch, ensuring robustness even when comparing against empty support hypotheses.

Formally, let $L$ be the number of characters in translation $t$. 
An error annotation $E$ consists of two index sets $E^{maj}, E^{min} \subseteq \{1, \dots, L\}$, representing the character positions of major and minor errors.
We represent this annotation not as discrete sets, but as a dense severity vector $\mathbf{v} \in \mathbb{R}^L$. 
For each character position $i \in \{1, \dots, L\}$, the vector element $\mathbf{v}^{(i)}$ represents the aggregated severity:
$\mathbf{v}^{(i)} = \beta \cdot \mathbb{I}[i \in E^{maj}] + \gamma \cdot \mathbb{I}[i \in E^{min}]$, 
where $\beta$ and $\gamma$ represent the penalties for major and minor errors, respectively.
The total error severity of an annotation is the $\ell_1$ norm of its vector, $\|\mathbf{v}\|_1 = \sum_{i=1}^L |\mathbf{v}^{(i)}|$. 

We measure the discrepancy between a candidate hypothesis $E_c$ (with vector $\mathbf{v}_c$) and a support hypothesis $E_s$ (with vector $\mathbf{v}_s$) using the $\ell_1$ distance, $\|\mathbf{v}_c - \mathbf{v}_s\|_1$.
Based on these vector representations, we define Soft Precision (\textsc{SoftP}) and Soft Recall (\textsc{SoftR}) as:
\begin{align}
    \textsc{SoftP}(E_c, E_s) &= 1 - \frac{\|\mathbf{v}_c - \mathbf{v}_s\|_1}{L + \|\mathbf{v}_c\|_1}, \nonumber \\
    \textsc{SoftR}(E_c, E_s) &= 1 - \frac{\|\mathbf{v}_c - \mathbf{v}_s\|_1}{L + \|\mathbf{v}_s\|_1}.\nonumber
    \label{def:softP&R}
\end{align}
Here, $L$ is included in the denominator for normalization. 
Finally, \textsc{SoftF1}$(E_c, E_s)$ is computed as the harmonic mean of \textsc{SoftP}$(E_c, E_s)$ and \textsc{SoftR}$(E_c, E_s)$.

Unlike the standard $\textsc{F1}(\cdot,\cdot)$, $\textsc{SoftF1}(\cdot,\cdot)$ returns a continuous value reflecting the degree of mismatch. 
Crucially, even if the support hypothesis is empty ($\mathbf{v}_s = \mathbf{0}$), a candidate with a small error vector $\mathbf{v}_c$ will yield a high utility score closer to $1$, rather than collapsing to $0$ (see proof in Appendix A.2). 
\rev{Furthermore, a synthetic meta-evaluation on human annotations (Appendix A.3) confirms that $\textsc{SoftF1}(\cdot,\cdot)$ faithfully tracks fine-grained quality changes that $\textsc{F1}(\cdot,\cdot)$ misses in exactly this regime.
This property enables MBR decoding to meaningfully rank candidates even when the consensus among support hypotheses is that the translation is error-free.}

\section{Experiments on MBR Decoding}
\label{sec:exp}

\subsection{Experimental Setup}
\label{sec:exp_setup}

\paragraph{Datasets and Models}
We used the MQM annotations from the WMT24 Metrics Shared Task as our test set, which covers three translation directions: English$\rightarrow$German, English$\rightarrow$Spanish, and Japanese$\rightarrow$Chinese. 
To perform zero-shot, \textbf{reference-free} ESD using a GEMBA-MQM–style prompt (\citealp{gemba_mqm}), 
we prompted four popular LLMs, i.e., 
\code{Llama-3.3-70B-Inst} (\citealp{llama3}), 
\code{Gemma-2-27b-it} (\citealp{gemma2}), 
\code{Phi-4} (\citealp{phi4}), and \code{Mistral-Large-Inst}\footnote{\url{https://huggingface.co/mistralai/Mistral-Large-Instruct-2407}}. 
We verified the training data cutoff or release dates for all models to ensure no overlap with the test set, thereby mitigating potential data leakage risks  (\citealp{llama3}; \citealp{gemma2}; \citealp{phi4}).

\paragraph*{Hypothesis Generation and Decision}
For hypothesis generation, the model generates $N \in \{16, 64, 256\}$ hypotheses for each input,
with Top-$K$ ($K=10$) sampling (\citealp{topk}) and a temperature of $2.0$.
This high temperature promotes diversity among candidate hypotheses, which is expected to indirectly elevate utility variance and thereby reduce utility estimation error (\citealp{mbr_properties2}).
For the decision phase, we compared the three MBR decoding methods introduced in \S\ref{sec:mbr_utility} against \rev{two baselines: MAP and Majority Voting. 
Majority Voting selects the candidate hypothesis that occurs most frequently in the candidate set $\mathcal{C}_{\text{ESD}}$.}
Following prior work (\citealp{mqm}; \citealp{metricx24}; \citealp{wmtqe23}), we merged ``critical'' errors into the ``major'' category and adopt their
recommended parameter values for Eq.~\ref{def:score}: $w_{\text{maj}}=-5$, $w_{\text{min}}=-1$, $\alpha=-25$, $\beta=1$ and $\gamma=0.5$.
We used guided generation (\citealp{guided_gen}) to ensure that each hypothesis adheres to the specified JSON format.
Furthermore, due to the presence of empty translations ($L=0$) in the WMT Metrics
Shared Task dataset that do not conform to the ESD task definition, we added $1$ to the denominators of \textsc{SoftP}$(E_c, E_s)$ and \textsc{SoftR}$(E_c, E_s)$ in practice to prevent division by zero.

\paragraph{Evaluation}
Following the WMT Metrics Shared Task, we used SPA (\citealp{spa}) and ${\text{Acc}}_{\text{eq}}^{*}$ (\citealp{acc_eq}) for system- and sentence-level evaluation, respectively.
For the span level, we used $\textsc{SoftF1}(\cdot, E_y)$ as the metric and reported the average scores over test instances.
We selected $\textsc{SoftF1}(\cdot, E_y)$ over $\textsc{F1}(\cdot, E_y)$ \rev{because, as discussed in \S\ref{sec:softf1}, $\textsc{SoftF1}(\cdot,\cdot)$ has better mathematical properties; in particular, it avoids the zero-credit plateau of $\textsc{F1}(\cdot,\cdot)$ when handling empty or zero-overlap annotations. Our synthetic meta-evaluation (Appendix~\ref{sec:meta_eval}) further confirms that $\textsc{SoftF1}(\cdot, E_y)$ tracks fine-grained changes in annotation quality that $\textsc{F1}(\cdot, E_y)$ fails to register.}
For the SPA and ${\text{Acc}}_{\text{eq}}^{*}$, we employed the PERM-BOTH test (\citealp{perm_both}).
For $\textsc{SoftF1}$\footnote{\textsc{SoftF1} indicates the average of the return values by $\textsc{SoftF1}(\cdot, E_y)$ during evaluation across the entire dataset. Similar definitions apply to other functions.}, since it operates at the instance level, we used paired bootstrap resampling (\citealp{pbs}). 
We used \rev{both MAP and Majority Voting as baselines} for all statistical significance tests; \rev{a method is marked with \textsuperscript{†} only when it significantly outperforms both baselines.}
For all ESD methods, sentence-level scores were derived using Eq.~\ref{def:score},
system-level scores were calculated by averaging all sentence-level scores within that  MT system.
\begin{table*}[!ht]
\centering
\setlength\tabcolsep{3.5pt}
\scalebox{1.0}{
\begin{tabular}{l@{\hspace{1em}}ccccccccccc}
\toprule
~ & &\multicolumn{4}{c}{\textbf{\code{Llama-3.3-70B-Inst}}}  & \multicolumn{4}{c}{\textbf{\code{Gemma2-27b-it}}}\\
\cmidrule(r){3-6} \cmidrule(l){7-10}
$N$&Method& SPA & ${\text{Acc}}_{\text{eq}}^{*}$ & \textsc{SoftF1} & \textsc{F1} & SPA & ${\text{Acc}}_{\text{eq}}^{*}$ & \textsc{SoftF1} & \textsc{F1}\\
\midrule
1&Greedy Search&.820&.567&.922&.475&.816&.524&.900&.415\\
\midrule
\multirow{5}{*}{16} &MAP&.831&.565&.921&.537&.806&.529&.934&\textbf{.571}\\
 &\rev{Majority Voting}&.833&.564&.923&\textbf{.540}&.809&.520&.886&.454\\
\cdashline{2-10}
&MBR-\textsc{ScoreSim} &~~\textbf{.855}\textsuperscript{†}&~~\textbf{.573}\textsuperscript{†}&.906&.381&.802&~~\textbf{.552}\textsuperscript{†}&.890&.252\\
&MBR-\textsc{F1} &.839&.566&.918&.522&.810&.538&.924&.561\\
&MBR-\textsc{SoftF1} &.845&.565&\textbf{.929}&.510&~~\textbf{.832}\textsuperscript{†}&.531&\textbf{.935}&.545\\
\midrule
\multirow{5}{*}{64} &MAP &.831&.568&.922&.539&.797&.521&.933&.568\\
 &\rev{Majority Voting}&.826&.564&.925&\textbf{.543}&.809&.520&.891&.472\\
\cdashline{2-10}
&MBR-\textsc{ScoreSim} &~~\textbf{.849}\textsuperscript{†}&~~\textbf{.578}\textsuperscript{†}&.902&.347&.814&~~\textbf{.557}\textsuperscript{†}&.880&.197\\
&MBR-\textsc{F1} &.841&.568&.913&.529&.821&~~.537\textsuperscript{†}&.922&~~\textbf{.575}\textsuperscript{†}\\
&MBR-\textsc{SoftF1} &.843&.569&~~\textbf{.931}\textsuperscript{†}&.514&~~\textbf{.833}\textsuperscript{†}&~~.533\textsuperscript{†}&~~\textbf{.938}\textsuperscript{†}&.554\\
\midrule
\multirow{5}{*}{256} &MAP&.823&.568&.919&.531&.844&.519&.926&.545\\
 &\rev{Majority Voting}&.831&.560&.922&.534&.806&.529&.934&\textbf{.571}\\
\cdashline{2-10}
&MBR-\textsc{ScoreSim} &.843&~~\textbf{.579}\textsuperscript{†}&.898&.328&.815&~~\textbf{.560}\textsuperscript{†}&.873&.170\\
&MBR-\textsc{F1} &~~.844\textsuperscript{†}&.569&.923&\textbf{.532}&.814&~~.539\textsuperscript{†}&.916&~~\textbf{.579}\textsuperscript{†}&\\
&MBR-\textsc{SoftF1} &~~\textbf{.848}\textsuperscript{†}&~~.571\textsuperscript{†}&~\textbf{.932}\textsuperscript{†}&.513&\textbf{.846}&~~.532\textsuperscript{†}&~~\textbf{.938}\textsuperscript{†}&.553\\
\bottomrule
\end{tabular}}
\caption{Evaluation results for the WMT 2024 Metrics Shared Task, conducted with \code{Llama-3.3-70B-Inst} and \code{Gemma2-27b-it}. We employ SPA, ${\text{Acc}}_{\text{eq}}^{*}$, and $\textsc{SoftF1}$ as metrics at the system, sentence, and span levels, respectively. We report average scores across all translation directions. The best score for the same model with $N$ is in bold. \textsuperscript{†} indicates significantly better performance than \rev{both the MAP and Majority Voting baselines} with the same $N$ (\# of candidate hypotheses) on all translation directions ($p<0.05$).}
\label{tab:main_exp_1}
\end{table*}

\subsection{Experimental Results}\label{sec:results}
Drawing on the results presented in Table~\ref{tab:main_exp_1} and~\ref{tab:main_exp_2}, we compare the performance of our proposed MBR decoding instantiations against the \rev{MAP and Majority Voting baselines}.
Our key findings are detailed below.
\paragraph{Performance of MBR-\textsc{ScoreSim}}
At $N\ge64$, MBR-\textsc{ScoreSim} achieved significantly higher ${\text{Acc}}_{\text{eq}}^{*}$ scores compared to \rev{both MAP and Majority Voting}, but lacks consistent improvement on SPA and \textsc{SoftF1}.
This indicates a trade-off: MBR-\textsc{ScoreSim} is highly effective at the specific level it optimizes (sentence-level) but struggles to generalize to system or span-level improvements. 
Nevertheless, it remains valuable for applications that rely on sentence-level metrics, such as reranking for MT.

\paragraph{Performance of MBR-\textsc{F1}}
MBR-\textsc{F1} performed comparably to \rev{both MAP and Majority Voting} with only marginal fluctuations.
We attribute this stagnation to the empty-annotation defect in \textsc{F1}$(\cdot,\cdot)$ (\S\ref{sec:mbr_f1}), which hinders its effectiveness as an optimization objective and confirms the need for our soft alternative.

\paragraph{Performance of MBR-\textsc{SoftF1}}
At $N=256$, MBR-\textsc{SoftF1} consistently improves \textsc{SoftF1} over \rev{both MAP and Majority Voting}, with these improvements being statistically significant in every setup;
for SPA and ${\text{Acc}}_{\text{eq}}^{*}$, MBR-\textsc{SoftF1} typically matches or outperforms \rev{both baselines}, although these gains are not always statistically significant.
For example, with \code{Mistral-Large-Inst}, the SPA and ${\text{Acc}}_{\text{eq}}^{*}$ scores under MBR-\textsc{SoftF1} are numerically close to those of MAP \rev{and Majority Voting}, and the differences are not significant.
Furthermore, at $N=16$, MBR-\textsc{SoftF1} performed comparably to MAP with only marginal fluctuations,
highlighting a key limitation: MBR-\textsc{SoftF1} effectively trades computational cost for quality, requiring large sets of hypotheses ($N$). 
To address this resource dependency, we investigate MBR distillation in \S\ref{sec:exp_distill}.

\paragraph{Conclusion}
While MBR-\textsc{ScoreSim} and MBR-\textsc{F1} are constrained by optimization trade-offs or defects in their utility functions, MBR-\textsc{SoftF1} largely mitigates these issues when $N$ is sufficiently large.
Overall, these results support the conclusion that MBR decoding, instantiated with MBR-\textsc{SoftF1} and a sufficiently large candidate set, significantly improves span-level performance and generally matches or outperforms \rev{both the MAP and Majority Voting baselines} at system and sentence levels.
Finally, we note a limitation regarding span-level conclusions, which depend on $\textsc{SoftF1}(\cdot,\cdot)$ being a reliable span-level metric.
Since MBR-\textsc{SoftF1} was directly optimized by $\textsc{SoftF1}(\cdot,\cdot)$ as the utility function, the evaluation metric intrinsically favors this method, a phenomenon known as \emph{metric bias} (\citealp{metric_bias}).
\rev{A remedy validated in MT is to use an ensemble of multiple distinct metrics as the utility function, which has been shown to mitigate metric bias (\citealp{metric_bias}).
Lacking a neutral metric, we leave a more thorough investigation of this bias to future work.}
\begin{table}[!h]
\setlength\tabcolsep{3pt}
\centering
\small{
\begin{tabular}{ccccccc}
\toprule
Method& SPA & ${\text{Acc}}_{\text{eq}}^{*}$ & \textsc{SoftF1} & \textsc{F1}\\
\midrule
Model-\textsc{ScoreSim} &.843&.579&.898&.328\\
\lyu{Model-\textsc{F1}}&.844&.569&.923&.532\\
Model-\textsc{SoftF1} &.848&.571&.932&.513\\
\midrule
Oracle-\textsc{ScoreSim} &.925&.926&.943&.701\\
Oracle-\textsc{F1}&.906&.776&.970&.843\\
Oracle-\textsc{SoftF1} &.902&.845&.981&.811\\
\bottomrule
\end{tabular}}
\caption{
Performance upper bound of MBR identified through an oracle setup. Analysis is conducted using \code{Llama-3.3-70B-Inst} with $N=256$. Model-* indicates MBR based on model-generated support hypotheses, while Oracle-* indicates MBR based on the human annotation.}
\label{tab:oracle}
\end{table}

\begin{table*}[!t]
\centering
\setlength\tabcolsep{3.5pt}
\scalebox{1.0}{
\begin{tabular}{lccccccccccc}
\toprule
~ & &\multicolumn{4}{c}{\textbf{\code{
Phi-4}}}  & \multicolumn{4}{c}{\textbf{\code{Mistral-Large-Inst}}}\\
\cmidrule(r){3-6} \cmidrule(l){7-10}
$N$&Method& SPA & ${\text{Acc}}_{\text{eq}}^{*}$ & \textsc{SoftF1} & \textsc{F1} & SPA & ${\text{Acc}}_{\text{eq}}^{*}$ & \textsc{SoftF1} & \textsc{F1}\\
\midrule
1&Greedy Search&.825&.548&.792&.307&.815&.538&.774&.171\\
\midrule
\multirow{5}{*}{16} &MAP &.817&.553&.920&.533&.819&\textbf{.541}&.913&.463\\
 &\rev{Majority Voting}&.809&.557&.901&.528&.820&.540&.915&.467\\
\cdashline{2-10}
&MBR-\textsc{ScoreSim} &~~.837\textsuperscript{†}&\textbf{.554}&.893&.284&.825&.530&.865&.119\\
&MBR-\textsc{F1}&.822&.528&.925&~~\textbf{.563}\textsuperscript{†}&~~\textbf{.834}\textsuperscript{†}&.536&.885&.454\\
&MBR-\textsc{SoftF1} &~~\textbf{.847}\textsuperscript{†}&.535&~~\textbf{.936}\textsuperscript{†}&~~.552\textsuperscript{†}&.828&.524&~~\textbf{.930}\textsuperscript{†}&~~\textbf{.484}\textsuperscript{†}\\
\midrule
\multirow{5}{*}{64} &MAP &.822&.551&.922&.550&.825&.530&.914&.471\\
 &\rev{Majority Voting}&.825&.553&.924&.555&.820&.531&.917&.476\\
\cdashline{2-10}
&MBR-\textsc{ScoreSim} &~~.\textbf{852}\textsuperscript{†}&~~\textbf{.559}\textsuperscript{†}&.881&.238&\textbf{.832}&~~\textbf{.549}\textsuperscript{†}&.857&.107\\
&MBR-\textsc{F1} &.831&.530&.924&~~\textbf{.575}\textsuperscript{†}&.830&.536&.879&~~\textbf{.498}\textsuperscript{†}\\
&MBR-\textsc{SoftF1} &~~.848\textsuperscript{†}&.535&~~\textbf{.938}\textsuperscript{†}&.559&.828&.528&~~\textbf{.933}\textsuperscript{†}&~~.496\textsuperscript{†}\\
\midrule
\multirow{5}{*}{256} &MAP&.832&.545&.921&.549&.841&.525&.905&.427\\
 &\rev{Majority Voting}&.825&.549&.920&.545&.834&.522&.911&.467\\
\cdashline{2-10}
&MBR-\textsc{ScoreSim} &.843&~~\textbf{.563}\textsuperscript{†}&.875&.222&.831&~~\textbf{.558}\textsuperscript{†}&.852&.104\\
&MBR-\textsc{F1} &.841&.536&.918&~~\textbf{.577}\textsuperscript{†}&.834&~~.538\textsuperscript{†}&.819&~~\textbf{.506}\textsuperscript{†}\\
&MBR-\textsc{SoftF1} &~~\textbf{.850}\textsuperscript{†}&.549&~~\textbf{.938}\textsuperscript{†}&~~.558\textsuperscript{†}&\textbf{.846}&.528&~~\textbf{.934}\textsuperscript{†}&~~.494\textsuperscript{†}\\
\bottomrule
\end{tabular}}
\caption{Evaluation results for the WMT 2024 Metrics Shared Task, conducted with \code{Phi-4} and \code{Mistral-Large-Inst}. The element definitions in this table are the same as those in Table \ref{tab:main_exp_1}.}
\label{tab:main_exp_2}
\end{table*}

\section{Analysis of MBR Decoding}
\label{sec:analysis}

\subsection{MBR Performance Upper Bound}
To understand the upper bound on MBR performance, we conducted an oracle experiment.
We evaluated the discrepancy between two MBR decoding strategies: (1) Oracle MBR, which computes utility against the human annotation (Eq.~\ref{def:mbr_esd})\footnote{We used the human annotation from the WMT Metrics Shared Task dataset, provisionally assuming that the probability of the human annotation for each $x$ is $1$.}, 
and (2) Standard MBR, which relies on model-generated support hypotheses (Eq.~\ref{def:mbr_esd_appro}).

As shown in Table \ref{tab:oracle}, Oracle MBR significantly outperforms the standard MBR (based on support hypotheses) across all metrics. This substantial gap indicates that the quality of the support hypotheses is a limiting factor, suggesting significant room for improvement in future work.
\rev{We expect this gap to narrow as the underlying model improves or as more effective hypothesis-generation strategies, such as epsilon sampling (\citealp{epsilon_sampling}), are adopted to produce higher-quality support hypotheses.}

\rev{We caution, however, that this oracle upper bound is likely optimistic.
It is computed against the single human annotation available for each instance and thus overfits to one annotator, whereas ESD is an inherently subjective task that exhibits substantial annotation variability (\citealp{label_variation}; \citealp{xcomet_calibrating}; \citealp{watchmen}).
A more realistic upper bound would account for multiple human annotations per instance, which we leave to future work.}

\subsection{MBR-\textsc{SoftF1} Scalability}
MBR decoding effectively leverages increased test-time computation to improve performance, a paradigm often referred to as \emph{test-time scaling} (\citealp{tts}).
Consequently, we investigate its scalability with respect to $N$.

\begin{table}[!h]
\setlength\tabcolsep{3pt}
\centering
\small{
\begin{tabular}{lccccccc}
\toprule
$N$&Method& SPA & ${\text{Acc}}_{\text{eq}}^{*}$ & \textsc{SoftF1} & \textsc{F1}\\
\midrule
\multirow{2}{*}{256} &MAP&.823&.568&.919&\textbf{.531}\\
&MBR-\textsc{SoftF1} &.848&\textbf{.571}&.932&.513\\
\midrule
\multirow{2}{*}{1024}&MAP&.831&.567&.916&.524\\
&MBR-\textsc{SoftF1} &\textbf{.849}&\textbf{.571}&\textbf{.934}&.513\\
\bottomrule
\end{tabular}}
\caption{Analysis of MBR-\textsc{SoftF1} scalability. Analysis is conducted using \code{Llama-3.3-70B-Inst}.}
\label{tab:N}
\end{table}

Due to computational constraints, we performed this analysis exclusively on the \code{Llama-3.3-70B-Inst}, increasing $N$ to $1024$. 
As presented in Table \ref{tab:N}, MBR-\textsc{SoftF1} shows negligible performance gains at $N=1024$ compared to $N=256$ on SPA, ${\text{Acc}}_{\text{eq}}^{*}$, and \textsc{SoftF1}.
We hypothesize that the model fails to generate sufficient diversity among candidates with the current temperature and Top-$K$ settings to benefit from the larger sample size. 

\subsection{Number and Proportion of Error Spans}
\label{sec:num_spans}

\begin{table}[!h]
\setlength\tabcolsep{6pt}
\centering
\small{
\begin{tabular}{cccc}
\toprule
Method& Major & Minor & $\frac{\text{Major}}{\text{Minor}}$\\
\midrule
Human&9.0K&15.5K&0.58\\
\midrule
MAP&18.9K&18.7K&1.01\\
MBR-\textsc{ScoreSim} &13.0K&59.6K&0.21\\
MBR-\textsc{F1}&20.5K&25.9K&0.79\\
MBR-\textsc{SoftF1} &13.1K&20.1K&0.65\\
\bottomrule
\end{tabular}}
\caption{Analysis of the error span distribution across different ESD annotators. Generative ESD metrics are based on \code{Llama-3.3-70B-Inst} with $N=256$.}
\label{tab:dist}
\end{table}

We analyzed the distribution of error spans in the outputs of human annotators versus generative ESD metrics. The results are summarized in Table \ref{tab:dist}.
We observe that MAP tends to over-generate error spans compared to human annotators. 
Furthermore, it exhibits a bias in severity proportion:  $\frac{\text{Major}}{\text{Minor}}$ for MAP ($1.01$) deviates significantly from the human reference ($0.58$).

Notably, although the utility functions for MBR-\textsc{F1} and MBR-\textsc{SoftF1} do not explicitly optimize for error counts, they effectively mitigate this proportional bias, bringing $\frac{\text{Major}}{\text{Minor}}$ much closer to human levels. 
Moreover, MBR-\textsc{SoftF1} produces absolute counts of major and minor errors that are closer to the human reference. 
Since the impact of this discrepancy in error distribution on downstream tasks remains an open question, we did not explore direct interventions in this study. 
However, a potential remedy for future work would be to incorporate a term related to error span counts into the utility function to explicitly control the distribution.

\subsection{Comparison with State-of-the-Art}
\begin{table}[!h]
\centering
\small{
\setlength\tabcolsep{2.2pt}
\begin{tabular}{ccccccc}
\toprule
Method& SPA & ${\text{Acc}}_{\text{eq}}^{*}$ & \textsc{SoftF1} & \textsc{F1}\\
\midrule
Llama-MAP&.823&.568&.919&\textbf{.531}\\
Llama-MBR-\textsc{SoftF1} &\textbf{.848}&.571&\textbf{.932}&.513\\
\midrule
xCOMET-ESD &.757&.553&.889&.302\\
\rev{xCOMET-ESD-Conf} &.720&.552&.911&.482\\
xCOMET-Reg &.844&\textbf{.581}&-&- \\
xCOMET-QE-ESD &.688&.541&.879&.289\\
\rev{xCOMET-QE-ESD-Conf} &.720&.532&.910&.450\\
xCOMET-QE-Reg &.825&.549&-&- \\
\bottomrule
\end{tabular}}
\caption{Comparison of MBR-\textsc{SoftF1} (with \code{Llama-3.3-70B-Inst} and $N=256$) vs. xCOMET. *-Reg and *-ESD denote regression (score-only) and token-classification modes, respectively.
xCOMET-QE-*  indicates a reference-free variant; others are reference-based.
\rev{*-Conf denotes a confidence-calibrated version of the corresponding *-ESD metric that discards predicted error spans whose confidence is below a threshold $\lambda$ (here $\lambda=0.7$); see Appendix B.}}
\label{tab:w/xcomet}
\end{table}
Table \ref{tab:w/xcomet} shows the comparison between generative metrics with MBR-\textsc{SoftF1} and the state-of-the-art ESD metric (\citealp{wmtm23}; \citealp{wmtm24}), xCOMET (\citealp{xcomet}) (\code{Unbabel/XCOMET-XXL}). 
According to SPA and \textsc{SoftF1}, Llama-MBR-\textsc{SoftF1} (i.e. MBR-\textsc{SoftF1} with \code{Llama-3.3-70B-Inst}) outperforms any variants of xCOMET, ranking lower than xCOMET-Reg only on ${\text{Acc}}_{\text{eq}}^{*}$.
\rev{The xCOMET-ESD variants tend to over-predict error spans, which weakens their span-level performance.
To partially compensate, we additionally report confidence-calibrated variants (xCOMET-ESD-Conf and xCOMET-QE-ESD-Conf) that discard predicted error spans whose confidence falls below a threshold $\lambda$ (we set $\lambda=0.7$).
This filtering substantially raises their span-level \textsc{SoftF1}, but at the expense of system- and sentence-level performance, and no single $\lambda$ uniformly improves over the unfiltered default; we present the complete analysis in Appendix B.}
While these results suggest that the generative ESD method with MBR decoding is competitive for the ESD task, it is worth noting that it entails much higher computational overhead than xCOMET.

\section{Experiment on MBR Distillation}
\label{sec:exp_distill}
Despite the significant performance gains demonstrated in \S\ref{sec:exp},
MBR decoding comes at a high computational cost. Specifically, it relies on generating and scoring a large set of hypotheses (i.e., a large $N$), which incurs substantial latency and compute overhead during inference.
To alleviate this bottleneck, we investigate \textbf{MBR distillation}, a technique designed to transfer the superior performance of MBR decoding into the model weights, thereby enabling comparable results using efficient greedy search.

Following prior work on MBR distillation (\citealp{mbr_distill1}; \citealp{mbr_distill2}; \citealp{mbr_distill3}), we employ Direct Preference Optimization (DPO) (\citealp{dpo}) to train the model to approximate MBR outputs.
DPO is a pairwise ranking optimization method typically used to align models with human preferences by maximizing the margin between a \emph{preferred} and a \emph{rejected} output.
We adapt DPO for MBR distillation by constructing a pairwise dataset where the candidate hypothesis with the highest utility is treated as the preferred output, and the one with the lowest utility serves as the rejected output.

\subsection{Experimental Setup}
\label{sec:exp_setup_distill}
\paragraph{Datasets and Model}
The training dataset was derived from $(s, t)$ pairs in the WMT20–22 Metrics Shared Task datasets (\citealp{wmtm20}; \citealp{wmtm21}; \citealp{wmtm22}).
We used the WMT24 Metrics Shared Task as the test set.
We used \code{Llama-3.3-70B-Inst} as the model.
To generate the training dataset, we employed the same MBR setup as in \S\ref{sec:exp}: $N=256$ with MBR-$\textsc{SoftF1}$.
We refer to the model distilled via DPO as \textbf{Distill-Greedy}, which employs standard greedy decoding at inference time.

\rev{\paragraph{Training Details}
We used the Lion optimizer (\citealp{lion}) with a learning rate of $10^{-7}$, a warmup step of $50$, a weight decay of $0.1$, a batch size of $256$, and train for a maximum of $5$ epochs.
We set the regularization term weight of DPO to $0.5$.
We trained the model in BF16 precision on a machine equipped with 8 NVIDIA H200 GPUs.
We used DeepSpeed (\citealp{deepspeed}) Zero3 for multi-GPU training.
Our training codebase is based on the TRL library (\citealp{trl}).
The dataset comprises a total of $110,699$ samples. We randomly split the data, reserving $10\%$ for the validation set and using the remainder for the training set. 
Upon completion of training, the checkpoint with the lowest loss on the validation set was selected as the final model.}

\paragraph{Evaluation}
The evaluation metrics for ESD are identical to those used in \S\ref{sec:exp}.

\subsection{Experimental Results}\label{sec:results_distill}
\begin{table}[!h]
\setlength\tabcolsep{3pt}
\centering
\small{
\begin{tabular}{cccccccc}
\toprule
N&Method& SPA & ${\text{Acc}}_{\text{eq}}^{*}$ & \textsc{SoftF1} & \textsc{F1}\\
\midrule
1&Greedy Search&.820&.567&.922&.475\\
\midrule
\multirow{2}{*}{256} &MAP&.823&.568&.919&.531\\
&MBR-\textsc{SoftF1} &.848&.571&.932&.513\\
\midrule
1&Distill-Greedy&\textbf{.857}&\textbf{.572}&\textbf{.938}&\textbf{.538}\\
\bottomrule
\end{tabular}}
\caption{Comparison of MBR-\textsc{SoftF1} (with $N=256$) and the distilled model (Distill-Greedy, with greedy search). Results other than Distill-Greedy are cited from Table~\ref{tab:main_exp_1}. Experiments are conducted using \code{Llama-3.3-70B-Inst} on the WMT24 Metrics Shared Task. Distill-Greedy denotes the model distilled from MBR-\textsc{SoftF1} outputs via DPO.}
\label{tab:distil}
\end{table}

As evidenced in Table~\ref{tab:distil}, Distill-Greedy achieves scores nearly identical to MBR-\textsc{SoftF1} on SPA, ${\text{Acc}}_{\text{eq}}^{*}$, and \textsc{SoftF1}.
This result confirms that our distilled model using greedy search can match the performance of full MBR decoding while eliminating the dependency on large $N$.
Given that the computational cost of the distilled model with greedy search is substantially lower than that of the pre-distillation model with MBR decoding, 
we conclude that MBR distillation offers a promising pathway toward realizing efficient, high-performance generative ESD methods.

\section{Conclusions and Future Work}
\label{sec:conclusions}
\rev{We} introduced MBR decoding for generative ESD models, marking the first application of reranking techniques to automatic MT metrics. 
\rev{Experimental} results show that our MBR decoding, when instantiated with MBR-\textsc{SoftF1} and applied with sufficiently large candidate sets, significantly improves span-level performance and generally matches or outperforms \rev{both the MAP and Majority Voting baselines} at system and sentence levels.
Furthermore, we established that MBR distillation allows a model decoded via \rev{effective greedy search to match the performance of MBR.}

\rev{Future work should first examine whether the proposed approach remains effective under more demanding evaluation conditions.
Recent work argues that MT metrics should be evaluated on source texts that are harder to translate and drawn from more diverse domains, since existing benchmarks may be insufficiently challenging for modern MT systems and metrics (\citealp{wmt25m,watchmen}).

A second direction is to revisit the design of ESD evaluation and utility functions.
Character-level overlap metrics, including our proposed \textsc{SoftF1}, they do not treat error-span boundary quality as an explicit evaluation dimension.
Future work should investigate whether explicit boundary modeling is necessary for ESD, and how span-level utilities should be adapted when the task prioritizes approximate localization, exact boundary recovery, or downstream diagnostic use.

Finally, reduce metric bias in MBR decoding is an important direction.
A promising remedy, validated in MT, is to use ensembles of heterogeneous utility functions rather than a single metric (\citealp{metric_bias}).
Combining span-level and sentence-level utilities may yield decoding decisions that generalize more reliably across evaluation criteria.}

\iftaclpubformat

\else
\fi

\bibliography{tacl2021}
\bibliographystyle{acl_natbib}

\iftaclpubformat

\onecolumn

\appendix

\section{$\textsc{F1}(\cdot,\cdot)$ vs. $\textsc{SoftF1}(\cdot,\cdot)$}
\label{sec:f1_softf1}

\rev{\subsection{$\textsc{F1}(\cdot,\cdot)$}
For each character position $i\in\{1,\ldots,L\}$, let $m^i=\mathbb{I}[i\in E^{maj}]$ and $n^i=\mathbb{I}[i\in E^{min}]$ indicate whether the character is covered by a major or minor error span. 
We write $(m_c^i,n_c^i)$ and $(m_s^i,n_s^i)$ for the candidate annotation $E_c$ and the support/reference annotation $E_s$, respectively. The character-level true-positive credit is $c^i=\max(\beta m_c^i m_s^i,\beta n_c^i n_s^i,\gamma m_c^i n_s^i,\gamma n_c^i m_s^i)$, where $\beta$ and $\gamma$ are the exact-severity and cross-severity credit weights, with $\beta>\gamma\ge0$. Let $C=\sum_{i=1}^L c^i$ and $S(E)=\sum_{i=1}^L \max(m^i,n^i)$. 
Precision and recall are defined as:
\[
\textsc{P}=
\begin{cases}
1, & S(E_c)=S(E_s)=0,\\
C/S(E_c), & S(E_c)>0,\\
0, & \text{otherwise,}
\end{cases}
\qquad
\textsc{R}=
\begin{cases}
1, & S(E_c)=S(E_s)=0,\\
C/S(E_s), & S(E_s)>0,\\
0, & \text{otherwise.}
\end{cases}
\]
Finally, $\textsc{F1}(E_c,E_s)=2\textsc{P}\textsc{R}/(\textsc{P}+\textsc{R})$ if $\textsc{P}+\textsc{R}>0$, and $\textsc{F1}(E_c,E_s)=0$ otherwise. 
Under this convention, the empty-empty match receives score $1$, while any non-empty comparison with zero true-positive credit receives score $0$.

\paragraph{A Defect in $\textsc{F1}(\cdot, \cdot)$}
The key limitation of $\textsc{F1}(\cdot,\cdot)$ is not restricted to empty annotations. Whenever $S(E_c)+S(E_s)>0$, the metric can be written as $\textsc{F1}(E_c,E_s)=2C/(S(E_c)+S(E_s))$, with value $0$ when $C=0$. Therefore, all zero-credit annotation pairs collapse to the same minimum value:
$C=0\ \text{and}\ S(E_c)+S(E_s)>0
\Longrightarrow
\textsc{F1}(E_c,E_s)=0$.
This includes at least three cases: $E_c\neq\emptyset,E_s=\emptyset$; $E_c=\emptyset,E_s\neq\emptyset$; and $E_c\neq\emptyset,E_s\neq\emptyset$ with no character-level credit. 
Thus, $\textsc{F1}(\cdot,\cdot)$ assigns the same score to a candidate that marks only one minor-error character and to a candidate that marks a large disjoint span, as long as both receive zero true-positive credit.

\subsection{$\textsc{SoftF1}(\cdot,\cdot)$: Non-empty vs.\ Empty Spans}
\label{sec:softf1_appendix}
If $E_c\neq\emptyset$ and $E_s=\emptyset$, then:
\begin{align*}
    \textsc{SoftP}(E_c, E_s) = 1 - \frac{\|\mathbf{v}_c\|_1}{L + \|\mathbf{v}_c\|_1} = \frac{L}{L + \|\mathbf{v}_c\|_1} > 0, ~~~
    \textsc{SoftR}(E_c, E_s) = 1 - \frac{\|\mathbf{v}_c\|_1}{L} > 0.
\end{align*}
which implies \(\textsc{SoftF1}(E_c, E_s=\emptyset)>0\).
This behavior differs from $\textsc{F1}(E_c, E_s=\emptyset)$ and highlights a desirable property of $\textsc{SoftF1}(E_c, E_s)$: when only one of \(E_c\) and \(E_s\) is empty, it does not simply return $0$ or $1$, but a decimal value between $0$ and $1$ that reflects the degree of the match.

\subsection{Synthetic Meta-Evaluation}
\label{sec:meta_eval}
The analysis above shows analytically that $\textsc{SoftF1}(\cdot,\cdot)$ avoids a defect of $\textsc{F1}(\cdot,\cdot)$; we now verify empirically that this yields a more faithful response to fine-grained quality changes on real annotations.

\paragraph{Setup}
On the entire WMT24 Metrics Shared Task, we perturb each human annotation $E_y$ with two operations.
\textbf{Add} introduces a spurious error: we pick a substring not covered by any error in $E_y$ and label a randomly sized portion of it (length drawn uniformly from $[1,\max(8,\ell)]$ for a chosen region of length $\ell$) as a major or minor error span.
\textbf{Delete} removes a randomly chosen error span from $E_y$ and thus applies only when $E_y$ is non-empty.
Each edit moves the annotation strictly farther from $E_y$, so a faithful metric should strictly decrease.
For each sample we apply up to six consecutive edits of a single operation, each building on the previously edited annotation.

\begin{table*}[!h]
\centering
\small{
\begin{tabular}{cccccc}
\toprule
Operation& Human Ref. & Edit Severity & Edits Num. &\textsc{DR-F1} & \textsc{DR-SoftF1}\\
\midrule
Add&Empty&Major&412K&0.173&1.000\\
Add&Empty&Minor&411K&0.173&1.000\\
\midrule
Add&Non-empty&Major&284K&1.000&1.000\\
Add&Non-empty&Minor&285K&1.000&1.000\\
\midrule
Delete&Non-empty&Major&311K&1.000&1.000\\
Delete&Non-empty&Minor&556K&1.000&1.000\\
\bottomrule
\end{tabular}}
\caption{Synthetic meta-evaluation on the WMT24 Metrics Shared Task. For each operation, edited-span severity, and reference type, ``Edits Num.'' is the number of perturbations applied, and the \textsc{DR-F1}/\textsc{DR-SoftF1} columns report the 
\emph{decline ratio}: 
the fraction of those edits for which the metric, computed against the original human annotation $E_y$, strictly decreases (higher is better).}
\label{tab:meta_eval}
\end{table*}

\paragraph{Results}
Table~\ref{tab:meta_eval} reports the results.
When $E_y$ is non-empty, both $\textsc{F1}(\cdot,\cdot)$ and $\textsc{SoftF1}(\cdot,\cdot)$ 
decline due to 
every perturbation, so both are reliable in this regime.
The two diverge sharply when $E_y$ is empty: $\textsc{F1}(\cdot,\cdot)$ declines with only $17.3\%$ of added errors, whereas $\textsc{SoftF1}(\cdot,\cdot)$ declines with all of them.
$\textsc{F1}(\cdot,\cdot)$ registers exactly one decline per editing sequence, so its decline ratio is the reciprocal of the average number of edits per sequence: $1/6\approx0.167$ for full six-edit sequences, with the observed $0.173$ slightly higher because some sequences terminate early, regardless of edit severity. 
In contrast, $\textsc{SoftF1}(\cdot,\cdot)$ strictly decreases with every added error.

\section{Confidence-Based Calibration of xCOMET}
\label{sec:xcomet_conf}
At inference time, the token-classification variants of xCOMET (xCOMET-ESD and xCOMET-QE-ESD) expose a hyperparameter $\lambda$ that filters out every predicted error span whose confidence score (in $[0.0, 1.0]$) falls below $\lambda$.
Adjusting $\lambda$ thus indirectly controls the number of predicted error spans, serving to prevent the model from over-predicting them.
We sweep $\lambda$ from $0.0$ to $1.0$ in increments of $0.1$.

\begin{table}[!ht]
\centering
\small{
\begin{tabular}{lcccccccccc}
\toprule
~ &\multicolumn{4}{c}{xCOMET-ESD}  & \multicolumn{4}{c}{xCOMET-QE-ESD}\\
\cmidrule(r){2-5} \cmidrule(l){6-9}
$\lambda$& SPA & ${\text{Acc}}_{\text{eq}}^{*}$ & \textsc{SoftF1} & \textsc{F1} & SPA & ${\text{Acc}}_{\text{eq}}^{*}$ & \textsc{SoftF1} & \textsc{F1}\\
\midrule
0.0 &.757&.553&.889&.302&.688&.541&.879&.289\\
0.1 &\textbf{.758}&\textbf{.554}&.888&.302&.687&\textbf{.542}&.879&.290\\
0.2 &\textbf{.758}&\textbf{.554}&.889&.302&.688&\textbf{.542}&.879&.290\\
0.3 &\textbf{.758}&\textbf{.554}&.889&.302&.687&.541&.879&.290\\
0.4 &.754&.553&.890&.309&.689&\textbf{.542}&.881&.293\\
0.5 &.753&.552&.898&.352&.689&\textbf{.542}&.886&.321\\
0.6 &.744&.542&.905&.419&.704&.537&.904&.384\\
0.7 &.720&.552&.911&.482&\textbf{.720}&.532&.910&.450\\
0.8 &.722&.526&.914&.523&.709&.530&.917&.522\\
0.9 &.703&.526&\textbf{.928}&.558&.679&.521&.922&.554\\
1.0 &.657&.518&\textbf{.928}&\textbf{.560}&.703&.519&\textbf{.925}&\textbf{.563}\\
\bottomrule
\end{tabular}}
\caption{Effect of the confidence-filtering threshold $\lambda$ on xCOMET-ESD and xCOMET-QE-ESD, evaluated on the WMT24 Metrics Shared Task (\code{Unbabel/XCOMET-XXL}, averaged over all translation directions). Predicted error spans with confidence below $\lambda$ are discarded; $\lambda=0.0$ corresponds to no filtering (the default). For each metric and model, the best score across $\lambda$ is in bold.}
\label{tab:xcomet_confidence}
\end{table}
As shown in Table~\ref{tab:xcomet_confidence}, SPA and ${\text{Acc}}_{\text{eq}}^{*}$ are negatively correlated with $\lambda$, whereas \textsc{SoftF1} is positively correlated with it.
Importantly, no value of $\lambda$ simultaneously surpasses the default ($\lambda=0.0$) across SPA, ${\text{Acc}}_{\text{eq}}^{*}$, and \textsc{SoftF1}.
We therefore adopt $\lambda=0.7$ as the representative xCOMET-*-Conf variant in the main text, because, relative to the unfiltered default ($\lambda=0.0$), it recovers a large portion of the span-level \textsc{SoftF1}/\textsc{F1} while still retaining a reasonable degree of SPA and ${\text{Acc}}_{\text{eq}}^{*}$.

We further investigate why \textsc{SoftF1} increases with $\lambda$.
Without filtering ($\lambda=0.0$), xCOMET-ESD predicts 48.4K major and 55.7K minor error spans, while xCOMET-QE-ESD predicts 40.7K major and 60.7K minor spans.
These counts are far larger than the 9.0K major and 15.5K minor spans produced by human annotators (Table~\ref{tab:dist}).
This pronounced discrepancy indicates that over-predicting error spans is a major cause of xCOMET's weak span-level performance, and it explains why simply increasing $\lambda$, thereby pruning low-confidence spans, substantially improves span-level scores.
We caution, however, that a larger $\lambda$ is not an effective way to improve xCOMET: the very filtering that boosts span-level metrics also causes a significant degradation in system- and sentence-level performance (SPA and ${\text{Acc}}_{\text{eq}}^{*}$).}

\fi
\end{document}